\documentclass[final]{cvsports}

\usepackage{times}
\usepackage{epsfig}
\usepackage{graphicx}
\usepackage{amsmath}
\usepackage{amssymb}

\usepackage{enumitem}
\usepackage[ruled]{algorithm2e}
\usepackage{array}
\usepackage{multirow}
\usepackage{color}
\usepackage{cases}
\usepackage[export]{adjustbox}
\usepackage[dvipsnames]{xcolor}
\usepackage{textpos}
\usepackage{comment}
\usepackage{textcomp}
\usepackage{listings}
\usepackage{bibentry}

\usepackage{rotating}

\usepackage{booktabs}
\newcommand{\B}{\bf}
\newcommand{\ant}[1]{\textcolor{anthored}{#1}}

\usepackage[pagebackref=true,breaklinks=true,colorlinks,bookmarks=false]{hyperref}

\def\cvprPaperID{8} 
\def\confYear{CVsports 2021}

\newcolumntype{C}[1]{>{\centering\let\newline\\\arraybackslash\hspace{0pt}}b{#1}}

\begin{document}

\title{SoccerNet-v2: A Dataset and Benchmarks for Holistic Understanding of Broadcast Soccer Videos}


\author{
\hspace{-0.03cm}Adrien Deli\`ege*\\
\hspace{-0.03cm}{\small University of Li\`ege}\\
\and
\hspace{-0.57cm} Anthony Cioppa*\\
\hspace{-0.57cm}{\small University of Li\`ege}\\
\and
\hspace{-0.57cm}Silvio Giancola*\\
\hspace{-0.57cm}{\small KAUST}\\
\and
\hspace{-0.57cm}Meisam J. Seikavandi*\\
\hspace{-0.57cm}{\small Aalborg University}\\
\and
\hspace{-0.57cm}Jacob V. Dueholm*\\
\hspace{-0.57cm}{\small Aalborg University}\\
\and
\hspace{-0.06cm}Kamal Nasrollahi\\
\hspace{-0.06cm}{\small Aalborg University, Milestone Systems}\\
\and
\hspace{-0.44cm}Bernard Ghanem\\
\hspace{-0.44cm}{\small KAUST}\\
\and
\hspace{-0.44cm}Thomas B. Moeslund\\
\hspace{-0.44cm}{\small Aalborg University}\\
\and
\hspace{-0.44cm}Marc Van Droogenbroeck\\
\hspace{-0.44cm}{\small University of Li\`ege}\\
}

\maketitle

\newcommand{\mysection}[1]{\vspace{2pt}\noindent\textbf{#1}}
\newcommand{\Table}[1]{Table~\ref{tab:#1}}
\newcommand{\Figure}[1]{Figure~\ref{fig:#1}}
\newcommand{\Equation}[1]{Equation~\eqref{eq:#1}}
\newcommand{\Equations}[2]{Equations \eqref{eq:#1} and \eqref{eq:#2}}
\newcommand{\Section}[1]{Section~\ref{sec:#1}}
\newcommand{\SoccerNet}{SoccerNet~\cite{Giancola_2018_CVPR_Workshops}\xspace}
\newcommand{\ActivityNet}{ActivityNet~\cite{caba2015activitynet}\xspace}

\newcommand{\TODO}[1]{\textcolor{red}{[TODO:#1]}}

\newcommand\blfootnote[1]{%
  \begingroup
  \renewcommand\thefootnote{}\footnote{#1}%
  \addtocounter{footnote}{-1}%
  \endgroup
}

\definecolor{myred}[a=.5]{RGB}{215,25,28} 
\definecolor{myorange}[a=.5]{RGB}{253,174,97}
\definecolor{anthoblue}[a=.5]{RGB}{31,119,180}
\definecolor{anthoorange}[a=.5]{RGB}{255,127,14}
\definecolor{anthogreen}[a=.5]{RGB}{0,150,0}
\definecolor{anthored}[a=.5]{RGB}{150,0,0}
\definecolor{anthobrown}[a=.5]{RGB}{153,76,0}
\definecolor{mygreen}[a=.5]{RGB}{166,217,106} 
\definecolor{mygray}[a=.5]{gray}{0.57}

\definecolor{newanthogreen}[a=.5]{RGB}{101,140,49}
\definecolor{newanthored}[a=.5]{RGB}{191,0,0}
\definecolor{newanthoblue}[a=.5]{RGB}{0,127,255}
\definecolor{newanthogray}[a=.5]{RGB}{76,76,76}

\definecolor{newanthoorangespotting}[a=.5]{RGB}{227,140,16}
\definecolor{newanthobluespotting}[a=.5]{RGB}{31,119,180}
\definecolor{newanthogreenspotting}[a=.5]{RGB}{44,160,44}

\definecolor{newanthoredreplay}[a=.5]{RGB}{183,27,27}
\definecolor{newanthopinkreplay}[a=.5]{RGB}{217,118,213}

\definecolor{newjacobblue}[a=.5]{RGB}{76,114,176}
\definecolor{newjacoborange}[a=.5]{RGB}{221,132,82}

\newcommand{\whitebox}{\hfill\textcolor{white}{\rule[1mm]{1.8mm}{2.8mm}}\hfill}
\newcommand{\redbox}{\hfill\textcolor{myred}{\rule[1mm]{1.8mm}{2.8mm}}\hfill}
\newcommand{\orangebox}{\hfill\textcolor{myorange}{\rule[1mm]{1.8mm}{2.8mm}}\hfill}
\newcommand{\greenbox}{\hfill\textcolor{mygreen}{\rule[1mm]{1.8mm}{2.8mm}}\hfill}
\newcommand{\graybox}{\hfill\textcolor{mygray}{\rule[1mm]{1.8mm}{2.8mm}}\hfill}
\newcommand{\BG}[1]{\textbf{{\color{red}[BG: #1]}}}

\blfootnote{\textbf{(*)} Equal contributions. More at \url{https://soccer-net.org/}. Contacts: adrien.deliege@uliege.be, anthony.cioppa@uliege.be, silvio.giancola@kaust.edu.sa, meisamjam@gmail.com, jvdu@create.aau.dk.}

\vspace{-1cm}

\begin{abstract}
    Understanding broadcast videos is a challenging task in computer vision, as it requires generic reasoning capabilities to appreciate the content offered by the video editing. 
    In this work, we propose SoccerNet-v2, a novel large-scale corpus of manual annotations for the \SoccerNet video dataset, along with open challenges to encourage more research in soccer understanding and broadcast production.
    Specifically, we release around 300k annotations within SoccerNet's 500 untrimmed broadcast soccer videos. We extend current tasks in the realm of soccer to include action spotting, camera shot segmentation with boundary detection, and we define a novel replay grounding task. For each task, we provide and discuss benchmark results, reproducible with our open-source adapted implementations of the most relevant works in the field.
    SoccerNet-v2 is presented to the broader research community to help push computer vision closer to automatic solutions for more general video understanding and production purposes.
\end{abstract}


\section{Introduction}
\label{sec:Intro}

Sports is a profitable entertainment sector, capping \$91 billion of annual market revenue over the last decade~\cite{GlobalSportsMarket}. \$15.6 billion alone came from the Big Five European Soccer Leagues (EPL, La Liga, Ligue 1, Bundesliga and Serie A) \cite{EuropeanFootballMarket,BigFiveMarket1,BigFiveMarket2}, with broadcasting and commercial activities being the main source of revenue for clubs~\cite{BroadcastingRevenue}. 
TV broadcasters seek to attract the attention and indulge the curiosity of an audience, as they understand the game and edit the broadcasts accordingly. In particular, they select the best camera shots focusing on actions or players, allowing for semantic game analysis, talent scouting and advertisement placement. With almost 10,000 games a year for the Big Five alone, and an estimated audience of 500M+ people at each World Cup~\cite{SoccerAudience}, automating the video editing process would have a broad impact on the other millions of games played in lower leagues across the world. Yet, it requires an understanding of the game and the broadcast production.

\begin{figure}
    \centering
    \includegraphics[width=\linewidth]{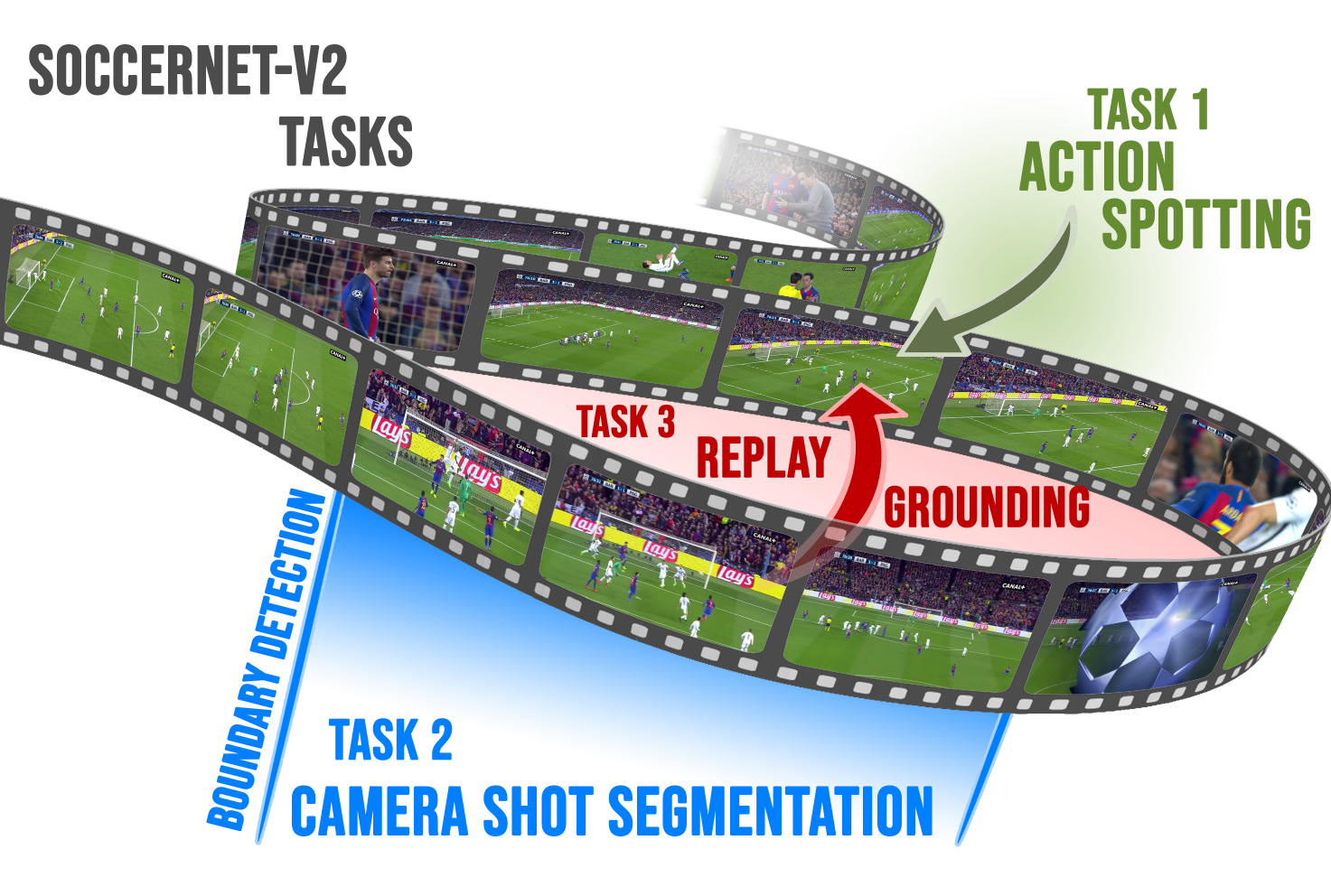}
    \caption{\textbf{SoccerNet-v2} constitutes the most inclusive dataset for soccer video understanding and production, with \texttildelow 300k annotations, 3 computer vision tasks and multiple benchmark results.}
    \label{fig:Pooling}
\end{figure}

Recent computer vision works on soccer broadcasts focused on low-level 
video understanding~\cite{moeslund2014computer}, \eg
localizing a field and its lines~\cite{Cioppa2018ABottom,farin2003robust,homayounfar2017sports}, 
detecting players~\cite{Cioppa_2019_CVPR_Workshops,yang2017robust}, 
their motion~\cite{felsen2017will,manafifard2017survey}, 
their pose~\cite{Bridgeman_2019_CVPR_Workshops, Zecha_2019_CVPR_Workshops}, 
their team~\cite{Istasse_2019_CVPR_Workshops}, 
the ball~\cite{Sarkar_2019_CVPR_Workshops,Theagarajan_2018_CVPR_Workshops},
or pass feasibility~\cite{Sangesa2020UsingPB}.
Understanding frame-wise information is useful to enhance the visual experience of sports viewers~\cite{Rematas_2018_CVPR} and to gather player statistics~\cite{thomas2017computer}, but it falls short of higher-level game understanding needed for automatic editing purposes (\eg camera shot selection, replay selection, and advertisement placement).


In this work, we propose a large-scale collection of manual annotations  
for holistic soccer video understanding and several benchmarks addressing automatic broadcast production tasks. In particular, we extend the previous \SoccerNet dataset with further tasks and annotations, and propose open challenges with public leaderboards. Specifically, we propose three tasks represented in Figure~\ref{fig:Pooling}:
\textbf{(i)} \textit{Action Spotting}, an extension from 3 to 17 action classes of SoccerNet's main task,
\textbf{(ii)} \textit{Camera Shot Understanding}, a temporal segmentation task for camera shots and a camera shot boundary detection task, 
and \textbf{(iii)} \textit{Replay Grounding}, a task of retrieving the replayed actions in the game. These tasks tackle three major aspects of broadcast soccer videos: action spotting addresses the understanding of the content of the game, camera shot segmentation and boundary detection deal with the video editing process, and replay grounding bridges those tasks by emphasizing salient actions, allowing for prominent moments retrieval. 

\mysection{Contributions.} We summarize our contributions as follows. 
\textbf{(i) Dataset.} We publicly release SoccerNet-v2, the largest corpus of manual annotations for broadcast soccer video understanding and production, comprising \texttildelow 300k annotations temporally anchored within SoccerNet's 764 hours of video. 
%
\textbf{(ii) Tasks.} We define the novel task of replay grounding 
and further expand the tasks of action spotting, camera shot segmentation and boundary detection, for a holistic understanding of content, editing, and production of broadcast soccer videos.
\textbf{(iii) Benchmarks.} We release reproducible benchmark results along with our code and public leaderboards to drive further research in the field.
   
\section{Related Work}
\label{sec:SOTA}

\mysection{Video understanding datasets.}
Many video datasets propose challenging tasks around 
human action understanding~\cite{gorelick2007actions,schuldt2004recognizing}, with applications in
movies~\cite{Kuehne11,marszalek2009actions,sigurdsson2016hollywood}, 
sports~\cite{KarpathyCVPR14,niebles2010modeling,rodriguez2008action}, 
cooking~\cite{damen2020epic,kuehne2014language,rohrbach2012database}, and large-scale generic video classification~\cite{abu2016youtube,kay2017kinetics,soomro2012ucf101}.
While early efforts focused on trimmed video classification, more recent datasets provide fine-grained annotations of longer videos at a 
temporal~\cite{caba2015activitynet,THUMOS14,sigurdsson2016hollywood,yeung2018every,zhao2019hacs} or spatio-temporal level~\cite{gu2018ava,mettes2016spot,rodriguez2008action,weinzaepfel2016human}. THUMOS14~\cite{THUMOS14} is the first benchmark for temporal activity localization, introducing 413 untrimmed videos, totalling 24 hours and 6k temporally anchored activities split into 20 classes, then extended to 65 classes in MultiTHUMOS~\cite{yeung2018every}.
ActivityNet~\cite{caba2015activitynet} gathers the first large-scale dataset for activity understanding, with 849 hours of untrimmed videos, temporally annotated with 30k anchored activities split into 200 classes. A yearly ActivityNet competition highlights a variety of tasks with hundreds of submissions~\cite{ghanem2018activitynet,ghanem2017activitynet}.
%
Some datasets consider videos at an atomic level, with fine-grained temporal annotations from short snippets of longer videos~\cite{gu2018ava,monfort2019moments,zhao2019hacs}. Multi-Moments in Time~\cite{monfort2019multi} provides 2M action labels for 1M short clips of 3s, classified into 313 classes.
Something-Something~\cite{goyal2017something} collects 100k videos annotated with 147 classes of daily human-object interactions.
Breakfast~\cite{kuehne2014language} and MPII-Cooking~2 \cite{rohrbach2016recognizing} provide annotations for individual steps of cooking activities. 
EPIC-KITCHENS~\cite{damen2020epic} scales up those approaches with 55 hours of cooking footage, annotated with around 40k action clips of 147 classes.

\mysection{Soccer-related datasets.}
\SoccerNet is the first large-scale soccer video dataset, with 500 games from major European leagues and 6k annotations. It provides complete games 
with a distribution faithful to official TV broadcasts, but 
 it only focuses on 3 action classes, 
 making the task too simplistic and of moderate interest.
SoccerDB~\cite{Jiang2020SoccerDB} adds 7 classes and player bounding boxes for half of SoccerNet's videos and 76 extra games. 
However, it misses a complete set of possible actions and editing annotations to allow for a full understanding of the production of TV broadcasts. 
Yu \etal~\cite{Yu2018Comprehensive} released a dataset with 222 halves of soccer matches with annotations of actions, shot transitions, and player bounding boxes. They have few annotations and do not carry out any experiment nor propose any task. 
Pappalardo \etal~\cite{Pappalardo2019Apublic} released a large-scale dataset of soccer events, localized in time and space. 
However, they focus on player and team statistics rather than video understanding, as they do not release any video. 
We address the limitations of these datasets by annotating all the interesting actions of the 500 SoccerNet games. Also, we provide valuable annotations for video editing, and we connect camera shots with actions to allow for salient moments retrieval. 


\mysection{Action spotting.} Giancola \etal~\cite{Giancola_2018_CVPR_Workshops} define the task of action spotting in SoccerNet as finding the anchors of soccer events in a video and provide baselines based on temporal pooling. Rongved \etal~\cite{rongved-ism2020} focus on applying a 3D ResNet directly to the video frames in a 5-second sliding window fashion. Vanderplaetse \etal~\cite{Vanderplaetse2020Improved} combine visual and audio features in a multimodal approach. Cioppa \etal~\cite{cioppa2020context} introduce a context-aware loss function to model the temporal context surrounding the actions. Similarly, Vats \etal~\cite{vats2020event} use a multi-tower CNN that accounts for the uncertainty of the action locations. Tomei \etal~\cite{tomei2020RMS} fine-tune a feature extractor and use a masking strategy to focus on the frames after the actions. 
We build upon those works to provide benchmark results on our extended action spotting task.

\mysection{Camera shot segmentation and boundary detection.}
Camera shot boundaries are typically detected by differences between frames, using pixels~\cite{Boreczky1996Comparison}, histograms~\cite{otsuji1994projection}, motion~\cite{zabih1995feature} or deep features~\cite{Abdulhussain2018MethodsAC}. 
In soccer, Hu \etal~\cite{Hu2007Enhanced} combine motion vectors and a filtration scheme to improve color-based methods. 
Lefèvre \etal~\cite{Lefvre2007EfficientAR} consider adaptive thresholds and features from a hue-saturation color space. 
Jackman~\cite{Jackman2019FootballSD} uses popular 2D and 3D CNNs but detects many false positives, as it appears difficult to efficiently process the temporal domain. Yet, these works are fine-tuned for only a few games. Regarding camera classification, Tong \etal~\cite{Tong2008Shot} first detect logos to select non-replay camera shots, further classified as long, medium, close-up or out-of-field views based on color and texture features. Conversely, Wang \etal~\cite{Wang2005SoccerRD} classify camera shots for the task of replay detection. Sarkar \etal~\cite{Sarkar2020Shot} classify each frame in the classes of~\cite{Tong2008Shot} based on field features and player dimensions. Kolekar \etal~\cite{Kolekar2015Bayesian} use audio features to detect exciting moments, further classified in camera shot classes for highlight generation.
In this paper, we offer a unified and enlarged corpus of annotations that allows for a thorough understanding of the video editing process.

\mysection{Replay grounding.}
In soccer, multiple works focus on detecting replays~\cite{Sarkar2020Shot, Wang2005SoccerRD,xu2011robust,yang2008statistical,zhao2006highlight},
using either logo transitions or slow-motion detection, but grounding the replays with their action in the broadcast has been mostly overlooked. 
Babaguchi \etal~\cite{babaguchi2000linking} tackle replay linking in American football but use a heuristic approach that can hardly generalize to other sports. 
Ouyng \etal~\cite{ouyang2005replay} introduce a video abstraction task to find similarities between multiple cameras in various sports videos, yet their method requires camera parameters and is tested on a restricted dataset. 
Replay grounding can be likened to action similarity retrieval, as in~\cite{hashemi2016view,junejo2008cross} for action recognition. Jain \etal~\cite{jain2020action} use a Siamese structure to compare the features of two actions, and Roy \etal~\cite{roy2018action} also quantify their similarity.
We propose a task of replay grounding to connect replay shots with salient moments of broadcast videos, which could find further uses in action retrieval and highlight production. 

\section{SoccerNet-v2 Dataset}
\label{sec:Dataset}

\begin{table*}[t]
\small
    \centering
    \caption{\textbf{Datasets.} Comparative overview of relevant datasets for action localization or spotting in videos. SoccerNet-v2 provides the second largest number of annotations and the largest in soccer. $^*$computed with the 116k annotations of the 200 fully annotated games.
    }
    \begin{tabular}{l||c|r|r|r|r|r|r}
\B Dataset                                   & \B Context & \B Duration & \B \#Actions & \B Classes & \B Density & \B Avg. events & \B Avg. video \\ 
& & \B (hrs) & & & \B (act./hr) & \B per class &\B length (sec)\\ 
\midrule 
\B THUMOS14~\cite{THUMOS14}                 & General  &                 24 &    6,363 & 20    & 260.4   & 318       & 209.2 \\
\B ActivityNet~\cite{caba2015activitynet}   & General  &                648 &   30,791 & 200   & 47.5    & 154       & 116.7    \\
\B Charades~\cite{sigurdsson2016hollywood}  & General  &                 82 &   66,500 & 157   & 811     & 424       & 30       \\
\B AVA~\cite{gu2018ava}                     & Movies   &              107.5 &  385,446 & 80    & 3,585   & 4,818     & 900      \\
\B HACS~\cite{zhao2019hacs}                 & Human    & \texttildelow2,166 &  139,000 & 200   & 64.2    & 695       & 156      \\
\B EPIC-Kitchen~\cite{damen2020epic}        & Cooking  &                 55 &   39,596 & 149   & 720     & 266       & 514.3    \\\midrule
\B \SoccerNet                               & Soccer   &                764 &    6,637 & 3     & 8.7     & 2,212     & 2750.4   \\
\B SoccerDB~\cite{Jiang2020SoccerDB}        & Soccer   &                669 &   37,715 & 11    & 56      & 3,428     & 14.1       \\
\B Yu \etal~\cite{Yu2018Comprehensive}        & Soccer   &                167 &   6,850 & 11    & 41      & 623     & 2708.1       \\
\B \ant{SoccerNet-v2 (actions)}             & Soccer   &                764 &  110,458 & 17    & 144     & 6,498     & 2750.4   \\
\B \ant{SoccerNet-v2 (cameras)}             & Soccer   &                306 &  158,493 & 13    & 381$^*$ & 8,976$^*$ & 2750.4   \\
\B \ant{SoccerNet-v2 (replays)}             & Soccer   &                764 &   32,932 & --    & 43      & --        & 2750.4   \\
    \end{tabular}
    \label{tab:DatasetsComparison}
\end{table*}

\begin{figure}
    \centering
    \includegraphics[width=\linewidth]{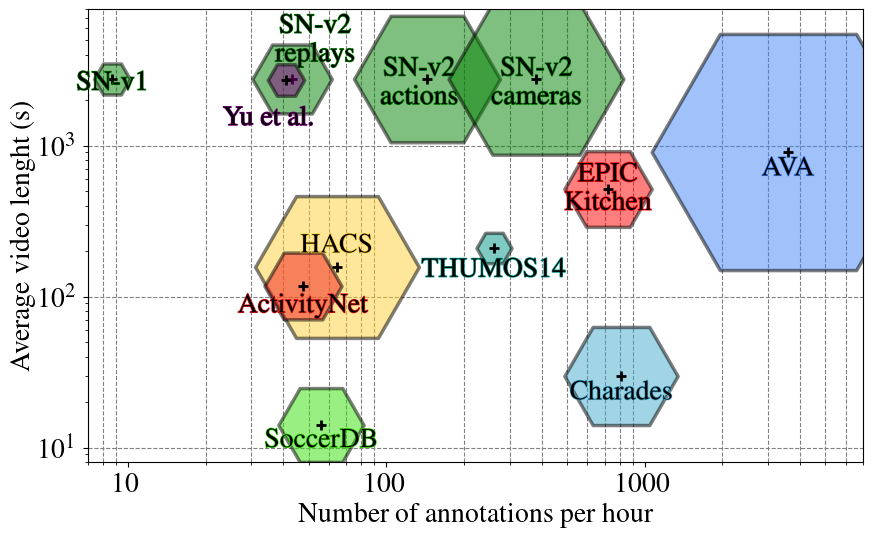}
    \caption{\textbf{Datasets comparison.} The areas of the tiles represent the number of annotations per dataset. SoccerNet-v2 (SN-v2) extends the initial \SoccerNet (SN-v1) with more annotations and tasks, and it focuses on untrimmed broadcast soccer videos.}
    \label{fig:datasets}
\end{figure}

\mysection{Overview.} Table~\ref{tab:DatasetsComparison} compares SoccerNet-v2 with the relevant video understanding datasets proposing localization tasks.  
SoccerNet-v2 stands out as one of the largest overall, and the largest for soccer videos by far. 
In particular, we manually annotated \texttildelow 300k timestamps, temporally anchored in the 764 hours of the 500 games of \SoccerNet. 
We center the vocabulary of our classes on the soccer game and soccer broadcast domains, hence it is well-defined and consistent across games. Such regularity makes SoccerNet-v2 the largest dataset in term of events instances per class, thus enabling deep supervised learning at scale.
%
As shown in Figure~\ref{fig:datasets}, SoccerNet-v2 provides the most dense annotations w.r.t. its soccer counterparts, and flirts with the largest fine-grained generic datasets in density and size. 

We hired 33 annotators for the annotation process, all frequent observers of soccer, for a total of \texttildelow 1600 hours of annotations. The quality of the annotations was validated by observing a large consensus between our annotators on identical games at the start and at the end of their annotation process. More details are provided in supplementary material. 
The annotations are divided in 3 categories: actions, camera shots, and replays, discussed hereafter. 

\begin{figure}
    \centering
    \includegraphics[width=1.0\linewidth]{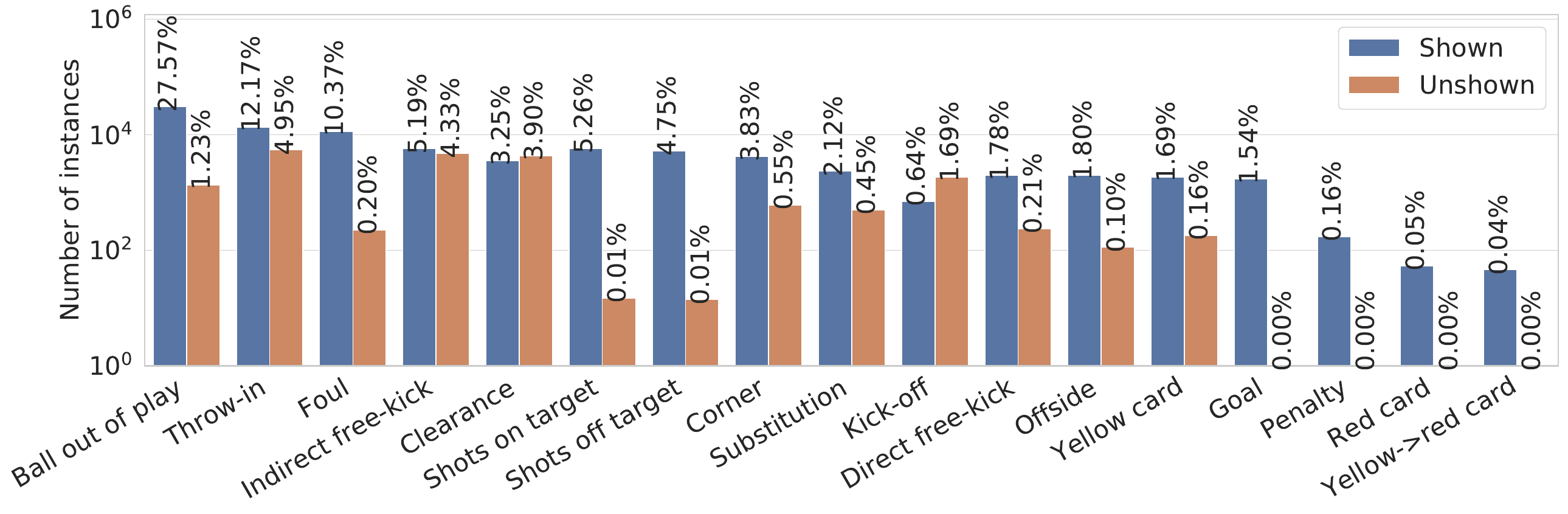}
    \caption{\textbf{SoccerNet-v2 actions.} Log-scale distribution of our \textcolor{newjacobblue}{\textbf{shown}} and \textcolor{newjacoborange}{\textbf{unshown}} actions among the 17 classes, and \textcolor{newanthogray}{\textbf{proportion}} that each class represents. The dataset is unbalanced, with some of the most important actions in the less abundant classes.
    }
    \label{fig:all-histos1}
\end{figure}

\mysection{Actions.} We identify 17 types of actions from the most important in soccer, listed in Figure~\ref{fig:all-histos1}. 
Following~\cite{Giancola_2018_CVPR_Workshops}, we annotate each action of the 500 games of SoccerNet with a single timestamp, defined by well-established soccer rules. For instance, for a corner, we annotate the last frame of the shot, \ie showing the last contact between the player's foot and the ball. We provide the annotation guidelines in supplementary material. 
In total, we annotate 110,458 actions, on average 221 actions per game, or 1 action every 25 seconds.
SoccerNet-v2 is a significant extension of the actions of \SoccerNet, with 16x more timestamps and 14 extra classes. 
We represent the distribution of the actions in Figure~\ref{fig:all-histos1}. 
The natural imbalance of the data corresponds to the distribution of real-life broadcasts, making SoccerNet-v2 valuable for generalization and industrial deployment.

\begin{figure}
    \centering
    \includegraphics[width=1.0\linewidth]{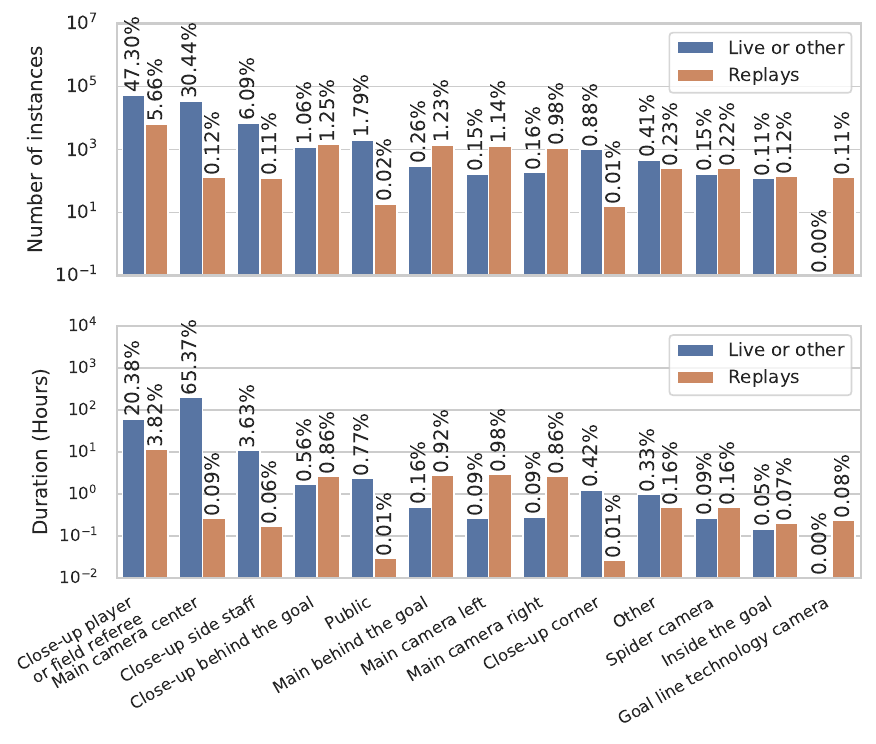}
    \caption{
    \textbf{Camera shots.} Log-scale distribution of our camera shot timestamps among the classes 
    in terms of instances (top) and video duration (bottom), separated in \textcolor{newjacoborange}{\textbf{replays}} and \textcolor{newjacobblue}{\textbf{live or other}} sequences, and \textcolor{newanthogray}{\textbf{percentage}} of timestamps that each bar represents.
    }
    \label{fig:all-histos2}
\end{figure}

Additionally, we enrich each timestamp with a novel binary visibility tag that states whether the associated action is \emph{shown} in the broadcast video or \emph{unshown}, in which case the action must be inferred by the viewer. For example, this happens when the producer shows a replay of a shot off target that lasts past the clearance shot of the goalkeeper: the viewer knows that the clearance has been made despite it was not shown on the TV broadcast. Spotting unshown actions is challenging because it requires a fine understanding of the game, beyond frame-based analysis, as it forces to consider the temporal context around the actions. 
We annotate the timestamps of unshown actions 
with the best possible temporal interpolation. 
They represent 18\% of the actions (see Figure~\ref{fig:all-histos1}), hence form a large set of actions whose spotting requires a sharp understanding of soccer.
Finally, to remain consistent with \SoccerNet, we annotate the team that performs each action as either \emph{home} or \emph{away}, but leave further analysis on that regard for future work.

\mysection{Cameras.} We annotate a total of 158,493 camera change timestamps, 116,687 of which are comprehensive for a subset of 200 games, the others delimiting replay shots in the remaining games (see hereafter). For the fully annotated games, this represents an average of 583 camera transitions per game, or 1 transition every 9 seconds. Those timestamps contain the type of camera shot that has been shown, among the most common 13 possibilities listed in Figure~\ref{fig:all-histos2}. 
We display their distribution in terms of number of occurrences and total duration. The class imbalance underpins a difficulty of this dataset, yet it represents a distribution consistent with broadcasts used in practical applications.

Besides, different types of transitions occur from one camera shot to the next, which we append to each timestamp. These can be abrupt changes between two cameras (71.4\%), fading transitions between the frames (14.2\%), or logo transitions (14.2\%). 
Logos constitute an unusual type of transition compared with abrupt or fading, which are common in videos in the wild or in movies, yet they are widely used in sports broadcasts. They pose an interesting camera shot detection challenge, as each logo is different and algorithms must adapt to a wide variety thereof. For logo and fading camera changes, we locate the timestamps as precisely as possible at the middle of the transition, while we annotate the last frame before an abrupt change.

Eventually, we indicate whether the camera shot happens live (86.7\%) with respect to the game, or shows a replay of an action (10.9\%), or another type of replay (2.4\%). The distribution in Figure~\ref{fig:all-histos2} provides per-class proportions of replay camera shots and groups other replays and live shots. 

\mysection{Replays.} For the 500 games of \SoccerNet, we bound each video shot showing a replay of an action with two timestamps, annotated in the same way as for the camera shot changes. For each replay shot, we refer the timestamp of the action replayed. When several replays of the same action are shown consecutively with different views, we annotate all the replay shots separately. This gives one replay shot per type of view, all of which are linked to the same action. In total, 32,932 replay shots are associated with their corresponding action, which represents an average of 66 replay shots per game, for an average replay shot duration of 6.8 seconds. 
Retrieving a replayed action is challenging because typically, 1 to 3 replays of the action are shown from different viewpoints hardly ever found in the original live broadcast video. This encourages a more general video understanding rather than an exact frame comparison. 

\section{Broadcast Video Understanding Tasks}

\begin{figure*}
    \centering
    \includegraphics[width=\linewidth]{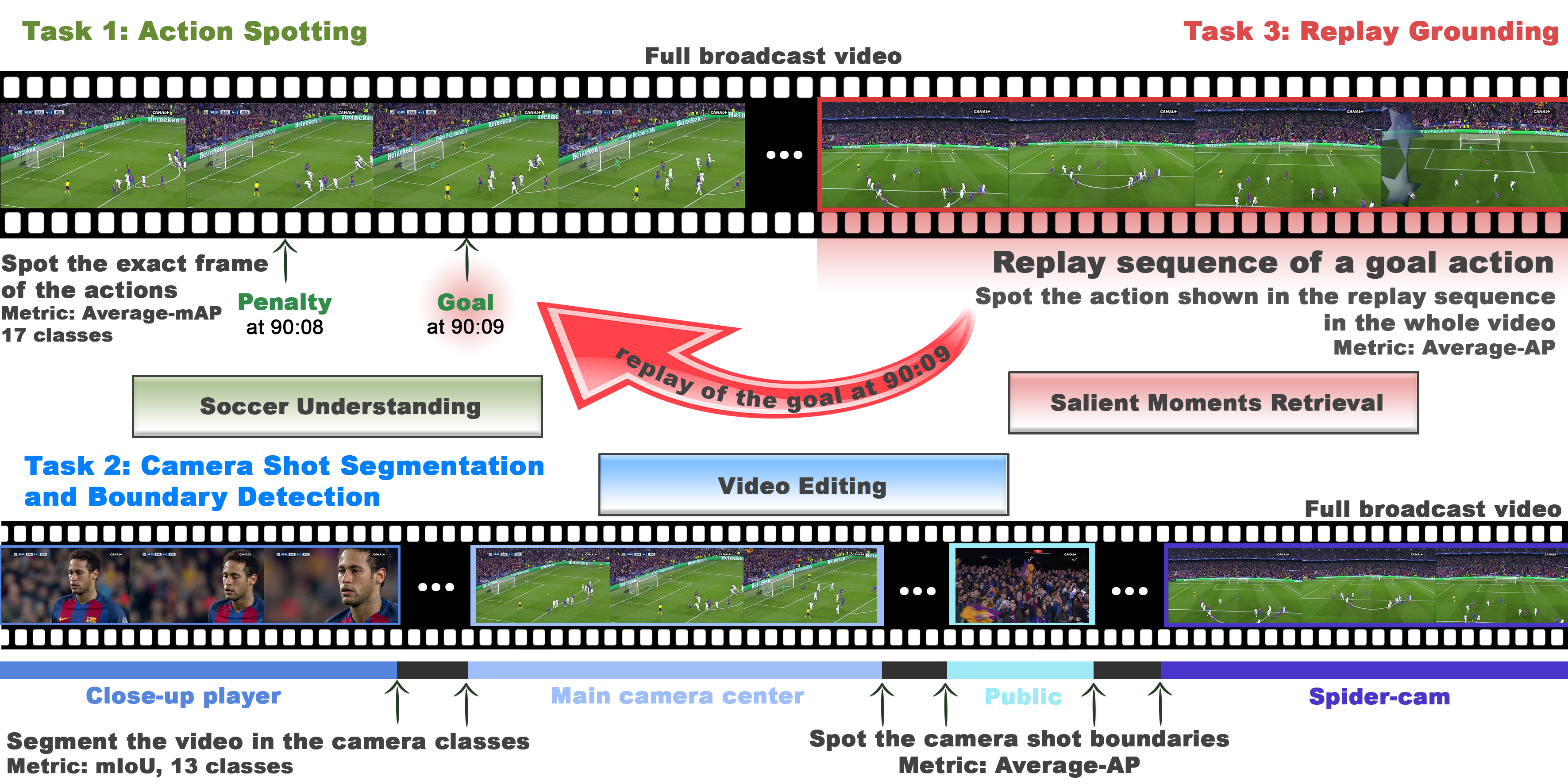}
    \caption{\textbf{Tasks overview.}
    We define a 17-class \textcolor{newanthogreen}{\textbf{action spotting}} task, a 13-class \textcolor{newanthoblue}{\textbf{camera shot segmentation}} and \textcolor{newanthoblue}{\textbf{boundary detection}} tasks, and a novel \textcolor{newanthored}{\textbf{replay grounding}} task, with their associated performance metrics. They respectively focus on \textcolor{newanthogreen}{\textbf{understanding the content}} of broadcast soccer games, addressing broadcast \textcolor{newanthoblue}{\textbf{video editing tasks}}, and \textcolor{newanthored}{\textbf{retrieving salient moments}} of the game.} 
    \label{fig:tasks}
\end{figure*}

We propose a comprehensive set of tasks to move computer vision towards a better understanding of broadcast soccer videos and alleviate the editing burden of video producers. More importantly, these tasks have broader implications as they can easily be transposed to other domains. This makes SoccerNet-v2 an ideal playground for developing novel ideas and implementing innovative solutions in the general field of video understanding.

In this work, we define three main tasks on SoccerNet-v2: action spotting, camera shot segmentation with boundary detection, and replay grounding, which are illustrated in Figure~\ref{fig:tasks}. They are further motivated and detailed hereafter.

\mysection{Action spotting.} 
In order to understand the salient actions of a broadcast soccer game, \SoccerNet introduces the task of action spotting, which consists in finding all the actions occurring in the videos. Beyond soccer understanding, this task addresses the more general problem of retrieving moments with a specific semantic meaning in long untrimmed videos. As such, we foresee moment spotting applications in \eg video surveillance or video indexing.

In this task, the actions are anchored with a single timestamp, contrary to the task of activity localization~\cite{caba2015activitynet}, where activities are delimited with start and stop timestamps. 
We assess the action spotting performance of an algorithm with the Average-mAP metric, defined as follows. A predicted action spot is positive if it falls within a given tolerance $\delta$ of a ground-truth timestamp from the same class. The Average Precision (AP) based on PR curves is computed then averaged over the classes (mAP), after what the Average-mAP is the AUC of the mAP computed at different tolerances $\delta$ 
ranging from 5 to 60 seconds.

\mysection{Camera shot segmentation and boundary detection.}
\label{subsec-Camera shot segmentation and boundary detection}
Selecting the proper camera at the right moment is the crucial task of the broadcast producer to trigger the strongest emotions on the viewer during a live game. Hence, identifying camera shots not only provides a better understanding of the editing process but is also a major step towards automating the broadcast production. This task naturally generalizes to any sports broadcasts but could also prove interesting for \eg cultural events or movies summarization.

Camera shot temporal segmentation consists in classifying each video frame among our 13 camera types and is evaluated with the 
 mIoU metric. 
 Concurrently, we define a task of camera shot boundary detection, where the objective is to find the timestamps of the transitions between the camera shots. For the evaluation, we use the spotting mAP metric with a single tolerance $\delta$ of 1 second as transitions 
are precisely localized and happen within short durations.
%

\mysection{Replay grounding.}
Our novel replay grounding task consists in retrieving the timestamp of the action shown in a given replay shot within the whole game. 
Grounding a replay with its action confers it an estimation of importance, which is otherwise difficult to assess. Derived applications may be further built on top of this task, \eg automatic highlight production, as the most replayed actions are usually the most relevant. 
Linking broadcast editing to meaningful content within the video not only bridges our previous tasks, but it can also be applied to any domain focusing on salient moments retrieval. 
We use the Average-AP to assess performances on this task, computed as described for the spotting task but without averaging over the classes. 
We choose this metric as replay grounding can be seen as class-independent action spotting conditioned by the replay sequence.
\section{Benchmark Results}
\label{sec:Exp}

\mysection{General comments.} \SoccerNet provides high and low quality videos of the 500 games. 
For easier experimentation, it also provides features from ResNet~\cite{He_2016_CVPR}, I3D~\cite{carreira2017quo} and C3D~\cite{tran2015learning} computed at 2 fps, further reduced with PCA to 512 dimensions.
Following~\cite{cioppa2020context,Giancola_2018_CVPR_Workshops}, in our experiments, we use the ResNet 512-dimensional frame features acting as compressed video representations as they yielded better results in early experiments. We adapt the most relevant existing methods to provide benchmark results on the \SoccerNet test set. We release our codes to reproduce them,
and we will host leaderboards on dedicated servers. 


\subsection{Action Spotting}

\mysection{Methods.}
We adapt or re-implement efficiently all the methods that released public code on \SoccerNet.

1. \emph{MaxPool} and \emph{NetVLAD}~\cite{Giancola_2018_CVPR_Workshops}.
Those models pool temporally the ResNet features before passing them through a classification layer. Non-overlapping segments of 20 seconds are classified as to whether they contain any action class. 
In testing, a sliding window of 20 seconds with a stride of 1 frame is used to infer an actionness score in time, reduced to an action spot using NMS. We consider the basic yet lightweight max pooling and a learnable NetVLAD pooling with 64 clusters. We re-implement the method 
based on the original code for a better scaling to 17 classes. 

2. \emph{AudioVid}~\cite{Vanderplaetse2020Improved}. The network uses NetVLAD to pool temporally 20-second chunks of ResNet features, as well as VGGish~\cite{Hershey2017CNN} synchronized audio features, subsampled at 2 fps. The two sets of features are temporally pooled, concatenated and fed to a classification module, as in~\cite{Giancola_2018_CVPR_Workshops}. 
Similarly, the spotting prediction is at the center of the video chunk.
We scaled the classification module to 17 classes.

3. \emph{CALF}~\cite{cioppa2020context}.
This network handles 2-minute chunks of ResNet features and is composed of a spatio-temporal features extractor, kept as is, a temporal segmentation module, which we adapt for 17 classes, and an action spotting module, adapted to output at most 15 predictions per chunk, classified in 17 classes. 
The segmentation module is trained with a context-aware loss having four context slicing hyperparameters per class.
Following~\cite{cioppa2020context}, we determine optimal values for them with a Bayesian optimization~\cite{BayesianOpt}. 
We re-implement the method and optimize the training strategy based on the existing code to achieve a decent training time.

\begin{figure}
    \centering
    \includegraphics[width=0.9\linewidth]{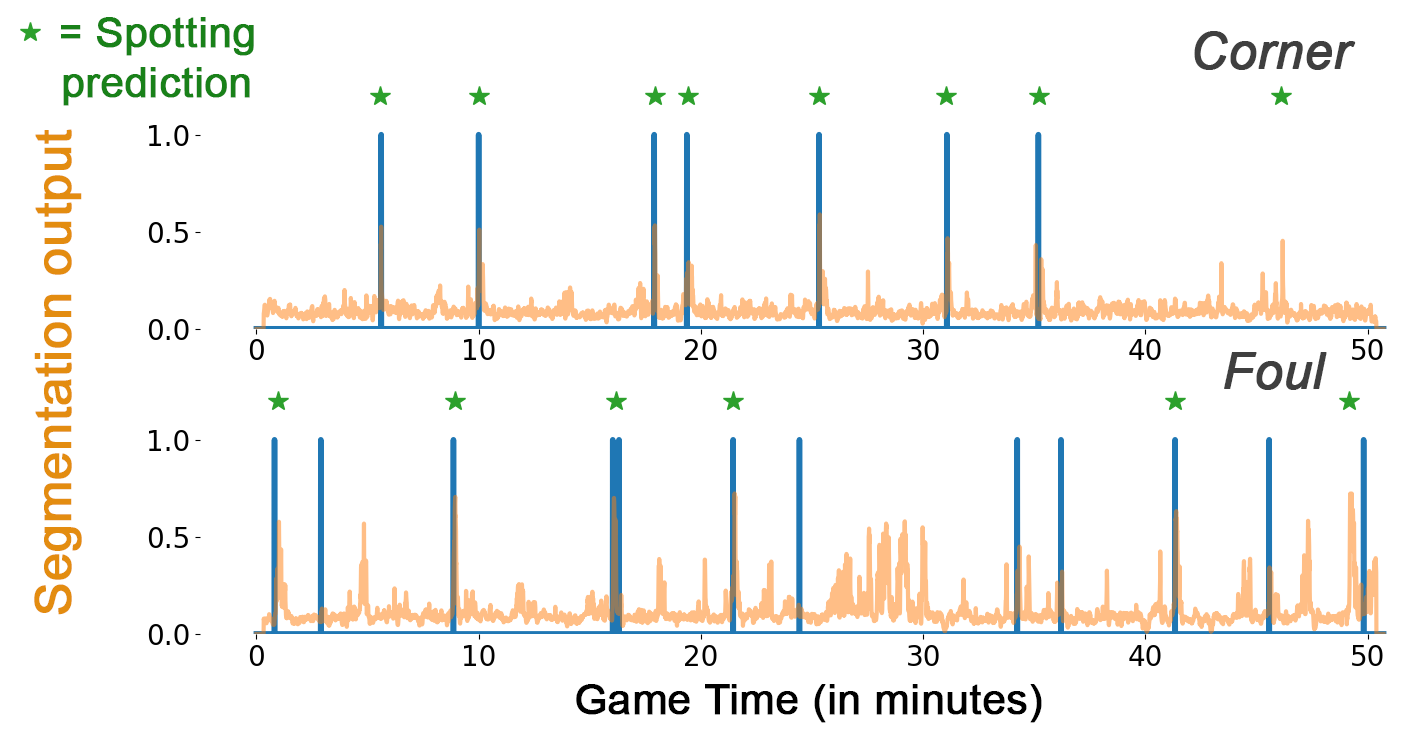}
    \caption{\textbf{Action spotting result} obtained from CALF adapted: \textcolor{newanthoorangespotting}{\textbf{temporal segmentation}}, \textcolor{newanthobluespotting}{\textbf{ground truth}}, and \textcolor{newanthogreenspotting}{\textbf{spotting predictions}}. The network performs well on corners with only one false positive, and moderately on fouls with a few false negatives.}
    \label{fig:result-action-spotting}
    \vspace{-10pt}
\end{figure}

\mysection{Results.}
We provide the leaderboard of our benchmark results for action spotting in Table~\ref{tab:ActionSpotting-long}. 
We further compute the performances on shown/unshown actions as the Average-mAP for predicted spots whose closest ground truth timestamp is a shown/unshown action.  
We show qualitative results obtained with CALF in Figure~\ref{fig:result-action-spotting}. 


The pooling approaches MaxPool and NetVLAD are not on par with the other methods on SoccerNet-v2. We believe the hard pruning with MaxPool has a restricted learning capacity, limited to a single fully connected layer. Similarly, NetVLAD may lag behind because of a non-optimal choice in the design of the spotting module, in particular the Non-Maximum Suppression that discards the results with confidence score below $0.5$. 
AudioVid prevails on the shown instances and on 5/17 actions classes. Injecting audio features appears to help with visible actions, as the sound is usually synchronized with the image. Also, it performs best on actions preceded or followed by the whistle of the referee, underlining the importance of audio features. Yet, the audio features appear less useful on unshown instances. 
%
CALF performs best globally, on the unshown instances and on most action classes. The context-aware loss focuses on the temporal context to spot the actions, which is useful for this task. This emphasizes the benefits of the temporal context surrounding the actions, that contains valuable information.

%

\begin{table*}[ht]
\scriptsize
    \caption{\textbf{Leaderboard for action spotting} (Average-mAP \%). Methods with codes publicly available were tested on SoccerNet-v2.} 
    \centering
    \setlength{\tabcolsep}{2pt}
    \resizebox{\linewidth}{!}{
    \begin{tabular}{l||c|c||c|c||c|c|c|c|c|c|c|c|c|c|c|c|c|c|c|c|c}
     &  \begin{turn}{90}SoccerNet-v1\end{turn}  &  \begin{turn}{90}SoccerNet-v2\end{turn} &  \begin{turn}{90}shown\end{turn}   &  \begin{turn}{90}unshown\end{turn}  & \begin{turn}{90} Ball out \end{turn} & \begin{turn}{90}Throw-in\end{turn} & \begin{turn}{90}Foul \end{turn} & \begin{turn}{90}Ind. free-kick \end{turn} & \begin{turn}{90}Clearance \end{turn} & \begin{turn}{90}Shots on tar. \end{turn} & \begin{turn}{90}Shots off tar. \end{turn} & \begin{turn}{90}Corner \end{turn} & \begin{turn}{90}Substitution \end{turn} & \begin{turn}{90}Kick-off \end{turn} & \begin{turn}{90}Yellow card \end{turn} & \begin{turn}{90}Offside \end{turn} & \begin{turn}{90}Dir. free-kick \end{turn} & \begin{turn}{90}Goal \end{turn} & \begin{turn}{90}Penalty \end{turn} & \begin{turn}{90}Yel.$\to$Red \end{turn} & \begin{turn}{90}Red card\end{turn} \\ 

       \midrule \midrule
Counts (test set)                   &   1369   & 22551 & 18641 & 3910 & 6460&	3809&	2414&	2283&	1631&	1175&	1058&	999&	579&	514&	431&	416&	382&	337&	41&	14&	8 \\ \midrule \midrule
MaxPool~\cite{Giancola_2018_CVPR_Workshops}   &   -   &  18.6 &  21.5 &  15.0 &  38.7 &  34.7 &  26.8 &  17.9 &  14.9 &  14.0 &  13.1 &  26.5 &  40.0 &  30.3 &  11.8 &   2.6 &  13.5 &  24.2 &   6.2 &  0.0 &\B0.9 \\ \midrule
NetVLAD~\cite{Giancola_2018_CVPR_Workshops}   &  49.7 &  31.4 &  34.3 &  23.3 &  47.4 &  42.4 &  32.0 &  16.7 &  32.7 &  21.3 &  19.7 &  55.1 &  51.7 &  45.7 &  33.2 &  14.6 &  33.6 &  54.9 &  32.3 &  0.0 &  0.0 \\ \midrule
AudioVid~\cite{Vanderplaetse2020Improved}     &  56.0 &  39.9 &\B43.0 &  23.3 &  54.3 &  50.0 &\B55.5 &  22.7 &  46.7 &  26.5 &  21.4 &  66.0 &\B54.0 &\B52.9 &  35.2 &  24.3 &\B46.7 &  69.7 &\B52.1 &  0.0 &  0.0 \\ \midrule
CALF~\cite{cioppa2020context}                 &  62.5 &\B40.7 &  42.1 &\B29.0 &\B63.9 &\B56.4 &  53.0 &\B41.5 &\B51.6 &\B26.6 &\B27.3 &\B71.8 &  47.3 &  37.2 &\B41.7 &\B25.7 &  43.5 &\B72.2 &  30.6 &\B0.7 &  0.7 \\ \bottomrule

\multicolumn{22}{l}{Other SoccerNet-v1 results but with no public code available: Rongved \etal~\cite{rongved-ism2020}: 32.0 ; Vats \etal~\cite{vats2020event}: 60.1 ; Tomei \etal~\cite{tomei2020RMS}: \textbf{75.1}.}


    \end{tabular}}
    \label{tab:ActionSpotting-long}
\end{table*}

\subsection{Camera Segmentation and Boundary Detection}

\mysection{Methods.} 
1. \emph{Basic model}. For our first baseline for the segmentation part, we train a basic model composed of 3 layers of 1D CNN with a kernel of 21 frames, hence aggregated in time, on top of ResNet features, and a MSE loss.

2. \emph{CALF (seg.)}~\cite{cioppa2020context}. We adapt CALF as it provides a segmentation module on top of a spatio-temporal features extractor. We replace its loss with the cross-entropy for easier experimentation and we focus on the segmentation by removing the spotting module. The number of parameters is reduced by a factor of 5 compared with the original model.

3. \emph{Content}~\cite{PySceneDetect}. For the boundary detection task, we test the popular scene detection library PySceneDetect. We use the Content option, that triggers a camera change when the difference between two consecutive frames exceeds a particular threshold value. This method is tested directly on the broadcast videos provided in \SoccerNet.

4. \emph{Histogram, Intensity}~\cite{scikitvideo}. We test two scene detection methods of the Scikit-Video library. The Histogram method reports a camera change when the intensity histogram difference between subsequent frames exceeds a given threshold~\cite{Otsuji93projection}. The Intensity method reports a camera change when variations in color and intensity between frames exceed a given threshold. Those methods are tested directly on the broadcast videos provided in \SoccerNet.

5. \emph{CALF (det.)}~\cite{cioppa2020context}. Since we can see the camera shot boundary detection as a spotting task, we recondition the best spotting method CALF by removing the segmentation module to focus on detection. Following a grid search optimization, we use 24-second input chunks of ResNet features and allow at most 9 detections per chunk.

\mysection{Results.} We provide a leaderboard of our benchmark results for these task in Table~\ref{tab:camera-shots-results}. 
We further compute the performances per transition type as the mAP for predicted spots grouped by the transition of their closest ground truth. 

Regarding the segmentation, even with 5x more parameters, the basic model trails behind CALF. Hence, simplistic architectures may not suffice for this task, and more sophisticated designs can rapidly boost performances. For the boundary detection, Histogram prevails, yet it ranks only third on fading transitions where the deep learning-based CALF is the best. The learning capacity of CALF may explain its performance consistency across transition types. 
Intensity, Content, and Histogram are intrinsically tailored for abrupt transitions. Intensity and Content are particularly bad on logos, while Histogram still performs well. 

  \begin{table}[t]
     \caption{\textbf{Leaderboard for Camera Shot Segmentation} (mIoU \%) and \textbf{Boundary Detection} (mAP \%).}
     \centering
     \setlength{\tabcolsep}{5pt}
     \resizebox{\linewidth}{!}{
     \begin{tabular}{c|c|c||c|c|c}
  & Camera & Bound. & \multicolumn{3}{c}{Transition} \\
  Method      & Seg.   & Det.  & Abrupt & Fading & Logo   \\ \midrule
  Basic model & 35.8 & - & - & - & - \\
  CALF~\cite{cioppa2020context} (seg.)& 47.3 & - & - & - & -\\
  CALF~\cite{cioppa2020context} (det.)& -    & 59.6 & 59.0 & \B58.0 & 61.8 \\
  Intensity~\cite{scikitvideo}        & -    & 64.0 & 74.3 & 57.2 & 28.5 \\
  Content~\cite{PySceneDetect}        & -    & 62.2 & 68.2 & 49.7 & 35.5 \\
  Histogram~\cite{scikitvideo}        & -    & \textbf{78.5} & \B83.2 & 54.1 & \B82.2\\ 
  \end{tabular}}
     \label{tab:camera-shots-results}
 \end{table}

\subsection{Replay Grounding}

\mysection{Methods.}
Given the novelty of this task, there is no off-the-shelf method available. 
We choose to adapt our optimized implementations of NetVLAD~\cite{Giancola_2018_CVPR_Workshops} and CALF~\cite{cioppa2020context} within a Siamese neural networks approach~\cite{Bromley1993Signature,Chicco2021Siamese,Koch2015SiameseNN}.

As input for the networks, we provide the ResNet features representations of a fixed-size video chunk and a replay shot. We either repeat or shorten the latter at both sides so that it has the same duration as the former. Ideally, for a chunk containing the action replayed (positive sample), the networks should output a high confidence score along with a localization prediction for spotting the action within the chunk. Otherwise (negative sample), they should only provide a low confidence score, and spotting predictions will be ignored. Negative samples are sampled either among chunks containing an action of the same class as the action replayed (hard negative), or among chunks randomly located within the whole video (random negative). 
The hard negatives ensure that the network learns to spot the correct actions without simply identifying their class, while the random negatives bring some diversity in the negative mining.

We test two sampling strategies. At each epoch, for each replay shot, we select: \emph{(S1)} only 1 sample: a positive with probability 0.5, or a hard or random negative each with probability 0.25; \emph{(S2)} 5 samples: 1 positive, 2 hard and 2 random negatives. For both S1 and S2, the positive is a chunk randomly shifted around the action timestamp.
The adaptations specific to each method are the following.

1. \emph{NetVLAD}~\cite{Giancola_2018_CVPR_Workshops}. We use NetVLAD to pool temporally the replay shot and the video chunk separately, but with shared weights. We compare the features obtained for the shot with those of the chunk through a cosine similarity loss, zeroed out when smaller than 0.4 to help the networks focus on improving their worst scores. In parallel, we feed the video features to a 2-layer MLP acting as spotting module to regress the spotting prediction within the chunk.   

\begin{figure}
    \centering
    \includegraphics[width=\linewidth]{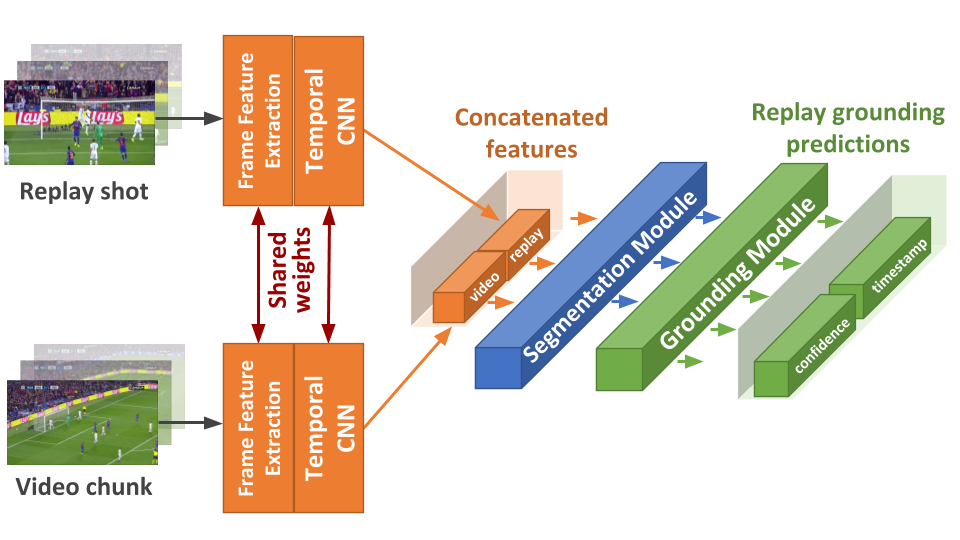}
    \caption{\textbf{Replay grounding pipeline} of our adaptation of CALF. In a Siamese network approach, the replay shot and the video chunk share a \textcolor{newanthoorangespotting}{\textbf{frame features extractor}}. Their features are concatenated and fed to the \textcolor{newanthobluespotting}{\textbf{segmentation module}}. The \textcolor{newanthogreenspotting}{\textbf{grounding module}} outputs a confidence score on the presence or absence of the action in the replay shot, and an action spotting prediction.}
    \label{fig:replay-grounding}
\end{figure}

2. \emph{CALF}~\cite{cioppa2020context}. We feed the replay shot and a video chunk to the shared frame feature extractor. Then, we concatenate the feature vectors along the temporal dimension, and give the resulting tensor to the remaining modules of the network. We set the number of classes to 1 in the segmentation module to provide per-frame insight. The spotting module outputs the confidence score on the presence of the replayed action in the chunk. We further set its number of detections to 1 as one action is replayed and might be spotted in the chunk. This architecture is represented in Figure~\ref{fig:replay-grounding}.

For these methods, at test time, we slice the video associated with the replay in chunks. We obtain at most one grounding prediction per chunk, all of which are kept when computing the Average-AP metric.

\begin{figure}
    \centering
    \includegraphics[width=\linewidth]{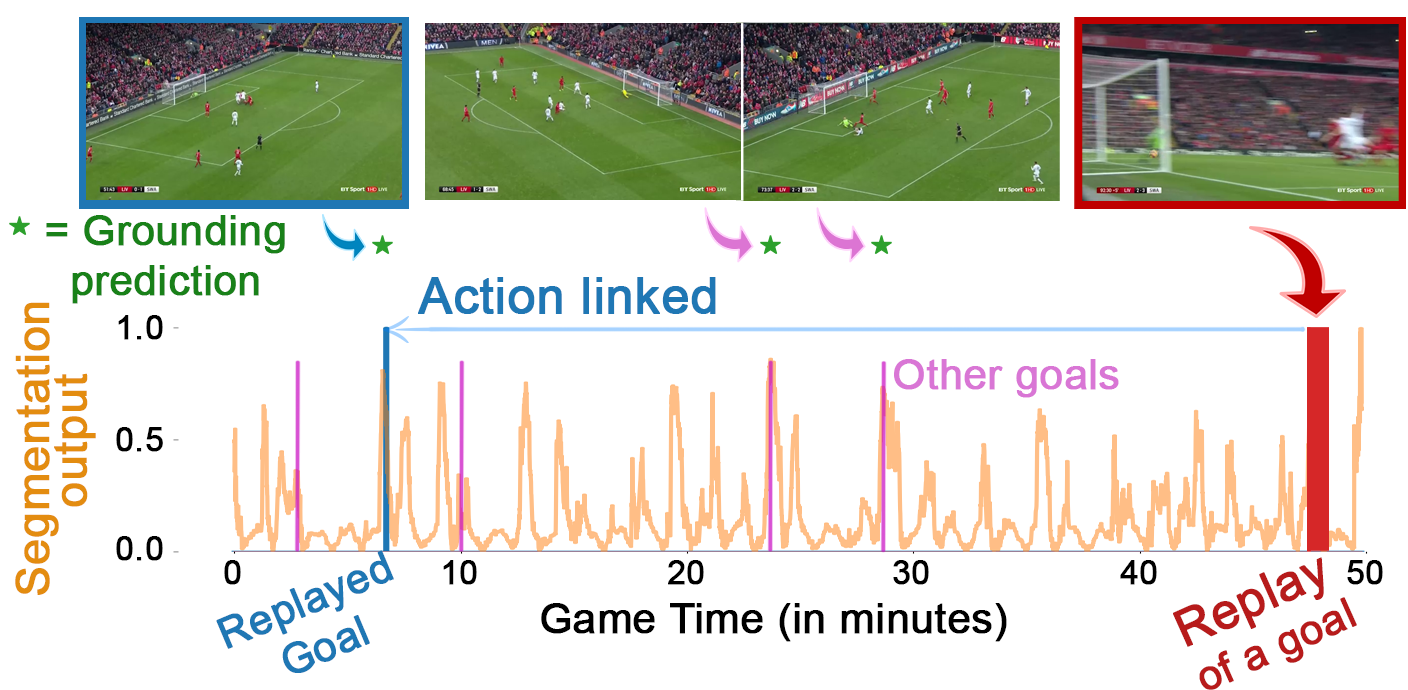}
    \caption{\textbf{Replay grounding result} of CALF adapted. We display the \textcolor{newanthoredreplay}{\textbf{replay shot}} of a goal, its \textcolor{newanthobluespotting}{\textbf{ground truth}} spot, the \textcolor{newanthopinkreplay}{\textbf{other goals}}, the temporal \textcolor{newanthoorangespotting}{\textbf{segmentation output}}, and the \textcolor{newanthogreenspotting}{\textbf{grounding predictions}}. The replayed goal is correctly spotted, two goals are rightly avoided, but two false positive predictions are also spotted, incidentally when other goals occurred. An insightful visualization can be appreciated in our \textbf{video in supplementary material}.}
    \label{fig:result-replay-grounding}
\end{figure}

\mysection{Results.}
The leaderboard providing our benchmark results for action spotting is given in Table~\ref{tab:Replay Grounding} for video chunks of different sizes. NetVLAD with S1 performs poorly, so no result is reported. 
Our adaptation of CALF achieves the best performance, with a chunk size of 60 seconds and with S2 as sampling strategy. Its demonstrated ability to aggregate the temporal context may explain this success. All the methods yield their best results with chunk sizes around 60 seconds, which presumably provides the most appropriate compromise between not enough and too much temporal context for an efficient replay grounding. An example of result from CALF is given in Figure~\ref{fig:result-replay-grounding}, showing that it can correctly learn to link a replay with its action without necessarily spotting all the actions of the same class. This underlines both the feasibility and the difficulty of our novel task. For a more relevant visualization experience, we invite the reader to consult our \textbf{video in supplementary material}.


\begin{table}[t]
\small
    \caption{\textbf{Leaderboard for replay grounding} (Average-AP \%), along with sampling strategy during training.}
    \centering
    \setlength{\tabcolsep}{5pt}
    \resizebox{\linewidth}{!}{
    \begin{tabular}{l|c|c|c|c|c|c|c}
    & \multicolumn{7}{c}{Video chunk size (seconds)} \\
Method     & 30 & 40 & 50 & 60 & 120 & 180 & 240    \\ \midrule
NetV.~\cite{Giancola_2018_CVPR_Workshops}+S2    & 23.9 & 22.9 &24.3 &  22.4 &  7.5 &  --  &  --  \\
CALF\cite{cioppa2020context}+S1                & 16.7 & 19.6 &  28.0 &32.3 & 32.0 & 26.9 & 22.0   \\
CALF\cite{cioppa2020context}+S2              &  8.2 & 14.7 &  28.9 &\B41.8 & 40.3 & 27.2 & 14.4
 \end{tabular}}
    \label{tab:Replay Grounding}
\end{table}

\section{Conclusion}
\label{sec:Conclusion}

We release SoccerNet-v2, the largest soccer-related set of annotations, anchored on top of the original \SoccerNet's 500 untrimmed broadcast games. 
With our \texttildelow 300k annotations, we further extend the tasks of action spotting, camera shot segmentation and boundary detection, and we define the novel task of replay grounding. 
We propose and discuss several benchmark results for all of them. In addition, we provide codes to reproduce our experiments, and we will host public leaderboards to drive research in this field. 
With SoccerNet-v2, we aim at pushing computer vision closer to automatic solutions for holistic broadcast soccer video understanding, and believe that it is the ideal dataset to explore new tasks and methods for more generic video understanding and production tasks. 

\mysection{Acknowledgments.} This work is supported by the DeepSport project of the Walloon Region, the FRIA, KAUST Office of Sponsored Research (award OSR-CRG2017-3405), and the Milestone Research Program at Aalborg University.

\clearpage

{\small
\bibliographystyle{ieee_fullname}
\bibliography{bibliography,biblio_CALF,biblio_SoccerNet}
}

\clearpage 

\newpage 
\section{Supplementary Material}

\subsection{Annotation Guidelines}

We provided our annotators with the following annotation guidelines to annotate the actions and the camera shots. 

\mysection{Actions.} Following the original \SoccerNet, we annotate each action with a single timestamp. These actions are illustrated in Figure~\ref{fig:action-types}, and their timestamps are defined as:

\begin{figure*}
    \centering
    \includegraphics[width=0.98\linewidth]{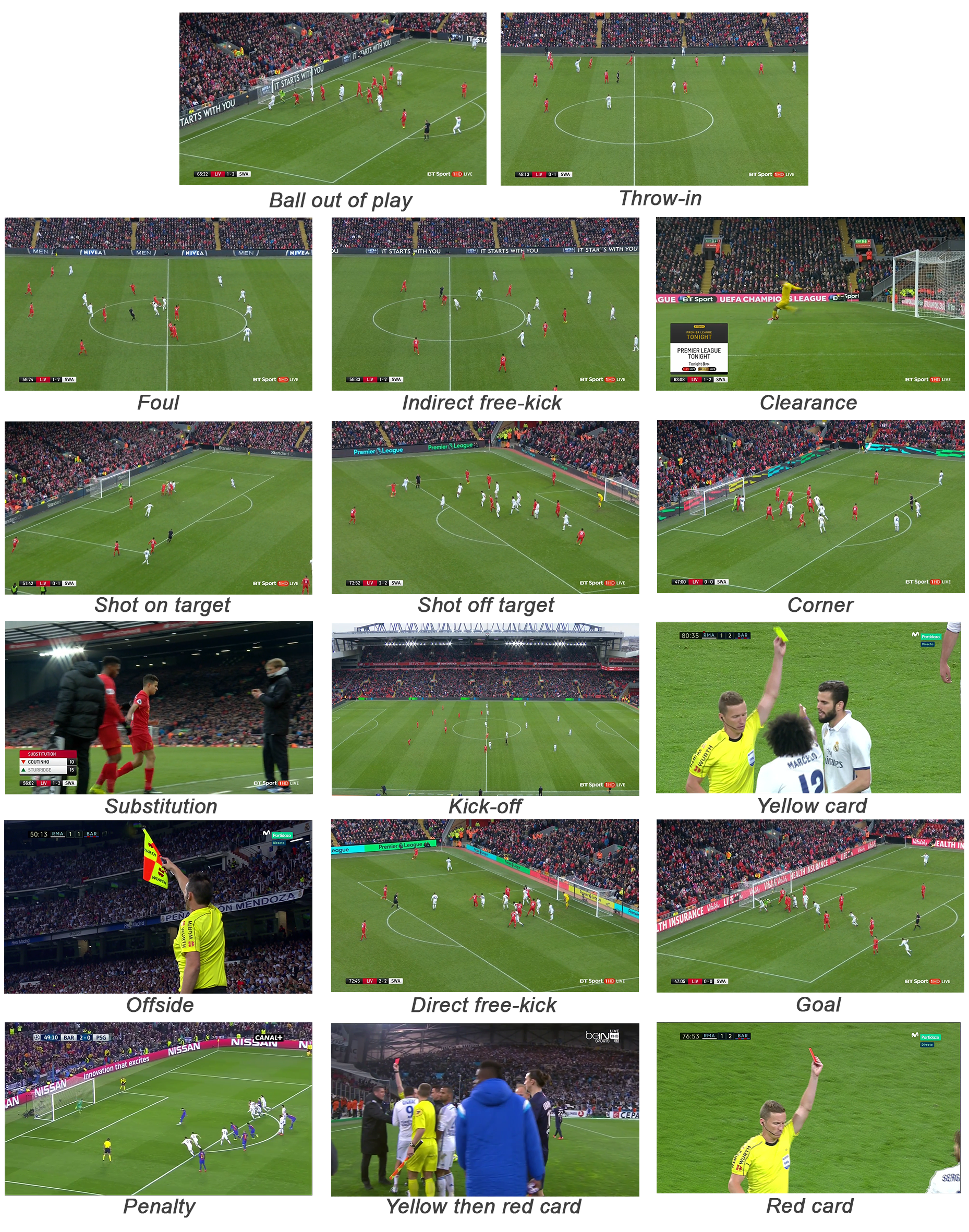}
    \caption{\textbf{Actions.} An example of each action identified in SoccerNet-v2.}
    \label{fig:action-types}
\end{figure*}

\begin{itemize}
    \item Ball out of play: Moment when the ball crosses one of the outer field lines.
    \item Throw-in: Moment when the player throws the ball  
    \item Foul: Moment when the foul is committed  
    \item Indirect free-kick: Moment when the player shoots, to resume the game after a foul, with no intention to score
    \item Clearance (goal-kick): Moment when the goalkeeper shoots
    \item Shots on target: Moment when the player shoots, with the intention to score, and the ball goes in the direction of the goal frame 
    \item Shots off target: Moment when the player shoots, with the intention to score, but the ball does not go in the direction of the goal frame
    \item Corner: Moment when the player shoots the corner
    \item Substitution: Moment when the replaced player crosses one of the outer field lines 
    \item Kick-off: Moment when, at the beginning of a half-time or after a goal, the two players in the central circle make the first pass
    \item Yellow card: Moment when the referee shows the player the yellow card
    \item Offside: Moment when the side referee raises his flag
    \item Direct free-kick: Moment when the player shoots, to resume the game after a foul, with the intention to score or if the other team forms a wall
    \item Goal: Moment when the ball crosses the line
    \item Penalty: Moment when the player shoots the penalty
    \item Yellow then red card: Moment when the referee shows the player the red card
    \item Red card: Moment when the referee shows the player the red card

\end{itemize}

\mysection{Camera shots.} We define the following 13 types of cameras, illustrated in Figure~\ref{fig:camera-types}:

\begin{figure*}
    \centering
    \includegraphics[width=\linewidth]{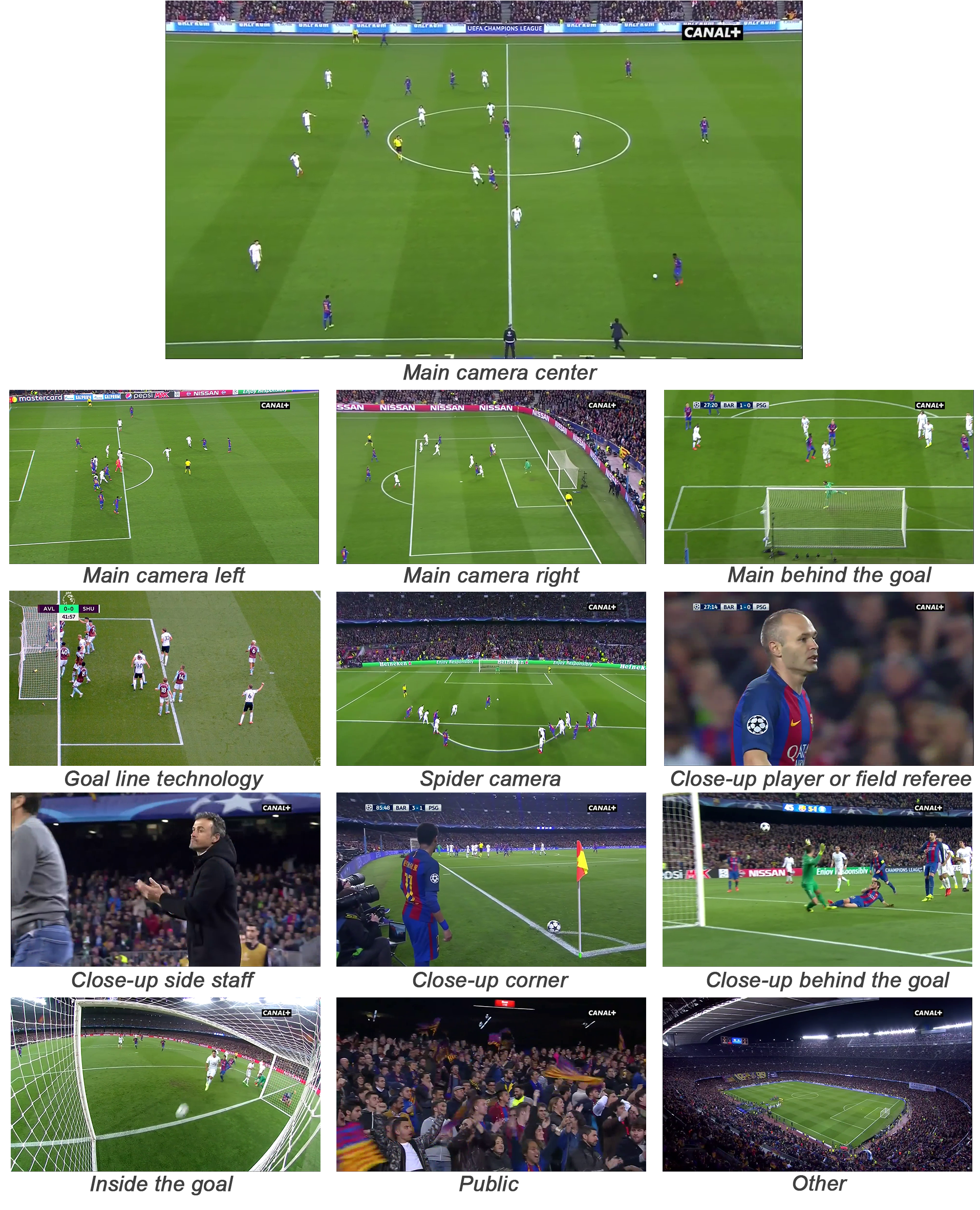}
    \caption{\textbf{Cameras.} An example of each camera shot identified in SoccerNet-v2.}
    \label{fig:camera-types}
\end{figure*}

\begin{itemize}
    \item Main camera center: Camera shown most of the time. It is placed high in the stadium and is centered on the middle field line. It films the players with a wide field of view, allowing an easy understanding of the game.
    \item Main camera left: Camera placed high in the stadium on the left side of the field. It is mostly used to allow for an easy overview of what is happening close to the left goal. It can also be used sometimes to show the right side of the field from a further perspective, mostly for an artistic effect. It is also sometimes called the 16-meter left camera.
    \item Main camera right: Counterpart of the main camera left but on the right side of the field.
    \item Main behind the goal: Camera placed behind the goal, either on a moving crane or in the stadium. It allows for a perpendicular field of view compared to the other main cameras.
    \item Goal line technology camera: Camera often placed next to the main camera left or right, but aligned with the goal line. It is used to check if the ball entirely crosses the line in contentious goal cases.
    \item Spider camera: Camera placed above the field and able to move freely in 3 dimensions thanks to long cables. It is often used in replays for a dynamic immersion in the action.
    \item Close-up player or field referee: Camera placed on ground-level, either fixed or at the shoulder of an operator, filming the players or the referees on the field with a narrower field of view. 
    \item Close-up side staff: Located similarly to close-up player cameras, films the reaction of the coaches and the staff outside the field. This also includes players on the bench or warming-up.
    \item Close-up corner: Camera often on the shoulder of an operator filming the player that shoots the corner.
    \item Close-up behind the goal: Camera either on the shoulder of an operator or fixed on the ground and filming the goalkeeper or the players from behind the goal.
    \item Inside the goal: Camera placed inside the goal that is sometimes shown during replays for an artistic effect.
    \item Public: Camera possibly located at different places in the stadium with the objective of filming the reaction of the public.
    \item Other: all other types of cameras that may not fit in the above definitions and that are most often used for artistic effects (\eg the helicopter camera or a split screen to show simultaneously two different games).
\end{itemize}

\subsection{Annotation Process}

\begin{figure*}
    \centering
    \includegraphics[width=\linewidth]{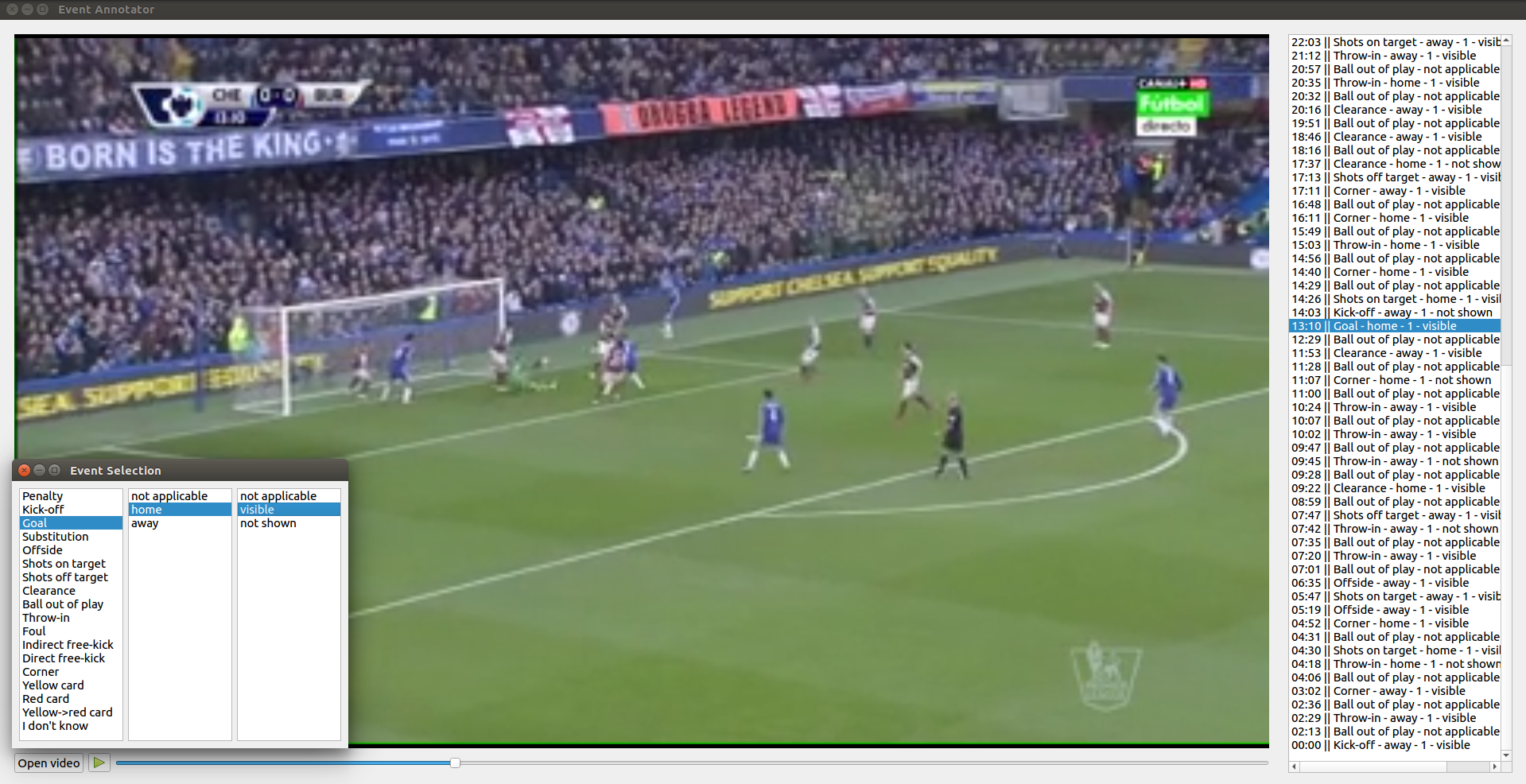}
    \caption{\textbf{Actions annotation tool.} When an action occurs, the annotator pauses the video to open the annotation menu (bottom left) and selects the action, the team that performs it, and whether it is shown or unshown in the video. The right column provides all the actions already annotated for that game, sorted chronologically.}
    \label{fig:annot-actions}
\end{figure*}

\begin{figure*}
    \centering
    \includegraphics[width=\linewidth]{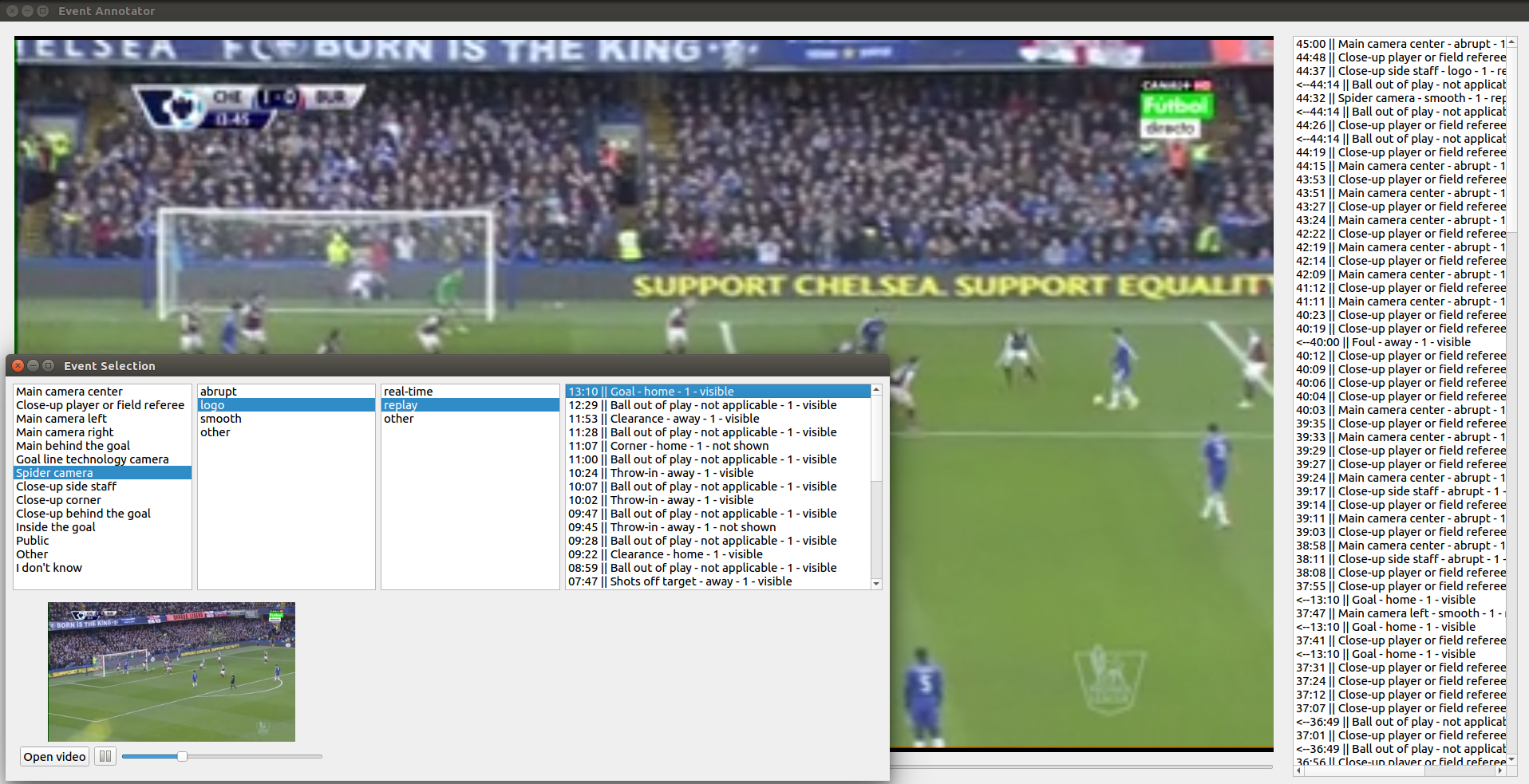}
    \caption{\textbf{Cameras annotation tool.} When a camera transition occurs (in this case, just before for a better visualization), the annotator pauses the video to open the annotation menu (bottom left) and selects the type of camera, the upcoming transition, and the real-time or replay characteristic of the current shot. In the case of a replay, as shown here, the annotator selects the action replayed in the last column, with the possibility to visualize a short clip around the action selected to ensure the correctness of the annotation. The large column on the right provides all the camera shots already annotated for that game, sorted chronologically.}
    \label{fig:annot-cameras}
\end{figure*}

We developed two tools for the annotations: the first for annotating the actions, shown in Figure~\ref{fig:annot-actions}, the second for the camera changes and replay grounding, shown in Figure~\ref{fig:annot-cameras}. For each video, a .json annotation file is created, which constitutes our annotations. The structure of the .json file is illustrated hereafter.

\lstset{basicstyle=\scriptsize\ttfamily}

\begin{lstlisting}[caption=Example of an action annotation in json.]
"UrlLocal": "path/to/game",
"annotations": [
    {
        "gameTime": "1 - 06:35",
        "label": "Offside",
        "position": "395728",
        "team": "away",
        "visibility": "visible"
    },
\end{lstlisting}

\begin{lstlisting}[caption=Example of a camera change annotation in json.]
"UrlLocal": "path/to/game",
"annotations": [
    {
        "change_type": "logo",
        "gameTime": "1 - 06:57",
        "label": "Main behind the goal",
        "link": {
            "half": "1",
            "label": "Offside",
            "position": "395728",
            "team": "away",
            "time": "06:35",
            "visibility": "visible"
        },
        "position": "417414",
        "replay": "replay"
    },
\end{lstlisting}

These tools were given to our 33 annotators, who are engineering students and soccer fans. Each annotator is attributed a given annotation task with detailed instructions and a set of matches to annotate. In case of doubt, they always have the possibility to contact us so that we control their work in ambiguous situations.

The total annotation time amounts to \texttildelow 1600 hours. Annotating all the actions of a single game takes \texttildelow 105 minutes; annotating all the camera changes requires \texttildelow 140 minutes per game, while only associating each replay shot of a game with its action takes \texttildelow 70 minutes.

\subsection{Human Level Performances}
Manually labeling events with timestamps raises the question of the sharpness of the annotations. In Charades~\cite{sigurdsson2016hollywood}, the average tIoU of human annotators on temporal boundaries is only of 72.5\%\footnote{Gunnar A. Sigurdsson, Olga Russakovsky, and Abhinav Gupta. What actions are needed for understanding human actions in videos? In \emph{IEEE International Conference on Computer Vision (ICCV)}, pages 2156-2165, October 2017.}, and 58.7\% on MultiTHUMOS~\cite{yeung2018every}.  Alwassel~\etal\footnote{Humam Alwassel, Fabian Caba Heilbron, Victor Escorcia, and Bernard Ghanem. Diagnosing error in temporal action detectors. In \emph{European Conference on Computer Vision (ECCV)}, pages 264-280, September 2018.} also observe some variability on ActivityNet~\cite{caba2015activitynet}, but note that a reasonable level of label noise still allows performance improvements and keeps the challenge relevant. 

Although all the annotations of our SoccerNet-v2 dataset are based on a set of well-defined rules, some uncertainty still resides in the timestamps. 
To quantify it, we determine an average human level performance on a common match shared across all the annotators as follows.
We assess the performance of an annotator against another by considering one as the predictor, the other as the ground truth. Then, we average the performances of an annotator against all the others to obtain his individual performance. Finally, we average the individual performances to obtain the human level performance. This yields an Average-mAP of 80.5\% for action spotting, a mIoU of 69.0\% for camera segmentation, and a mAP of 90.2\% for camera shot boundary detection. These metrics indicate that label noise is present but that current algorithms are still far from solving our tasks with a human-level cognition of soccer, as seen in Tables~\ref{tab:ActionSpotting-long},~\ref{tab:camera-shots-results},~\ref{tab:Replay Grounding} of the main paper.

\end{document}